\documentclass[11pt]{article}
\usepackage[utf8]{inputenc}
\usepackage[T1]{fontenc}
\usepackage{lmodern}

\usepackage{amsmath, amssymb, amsfonts}
\usepackage{multirow}
\usepackage{graphicx}
\usepackage{booktabs}
\usepackage{hyperref}
\usepackage{geometry}
\begin{document}

\title{Quality-Aware Modulation for Diffusion Transformers}

% \author{
% \small
% Luke Budny \\
% Carleton University \\
% \texttt{lukeswalks@hotmail.com}
% \and
% Yuhong Guo \\
% Carleton University \\
% \texttt{YuhongGuo@cunet.carleton.ca}
% \and
% Kevin Cheung \\
% Carleton University \\
% \texttt{Kevincheung@cunet.carleton.ca}
% }

\author{
Luke Budny, Yuhong Guo, Kevin Cheung \\
Carleton University \\
\texttt{LukeBudny@cmail.carleton.ca,\{YuhongGuo, KevinCheung\}@cunet.carleton.ca}
}

\date{}
\maketitle              % typeset the header of the contribution

\begin{abstract}
Modern text-to-image diffusion models, such as diffusion transformers (DiT), rely on timestep or prompt embeddings to modulate the strength of the denoising process in each timestep. While this modulation communicates the current noise level, it does not provide any quality-aware information, which can lead to generated images that are unaligned, visually inconsistent, and lacking in fidelity. In this paper, we propose the Quality Representation Module (QRM), a lightweight transformer module that learns a quality-aware representation based on existing model inputs, and produces a set of vectors $M_{qrm}$. These vectors adjust the adaptive LayerNorm modulation within the DiT transformer blocks, thereby injecting a quality-sensitive signal into the denoising parameters. The QRM introduces no significant changes to the sampling schedule or diffusion backbone. Experiments include ablations on QRM training losses and architectures, as well as empirical results demonstrating consistent image quality improvements over baseline DiT-based models.

\end{abstract}

\section{Introduction}
\label{sec:intro}

\begin{figure*}
    \centering
    \includegraphics[width=0.65\textwidth, keepaspectratio]{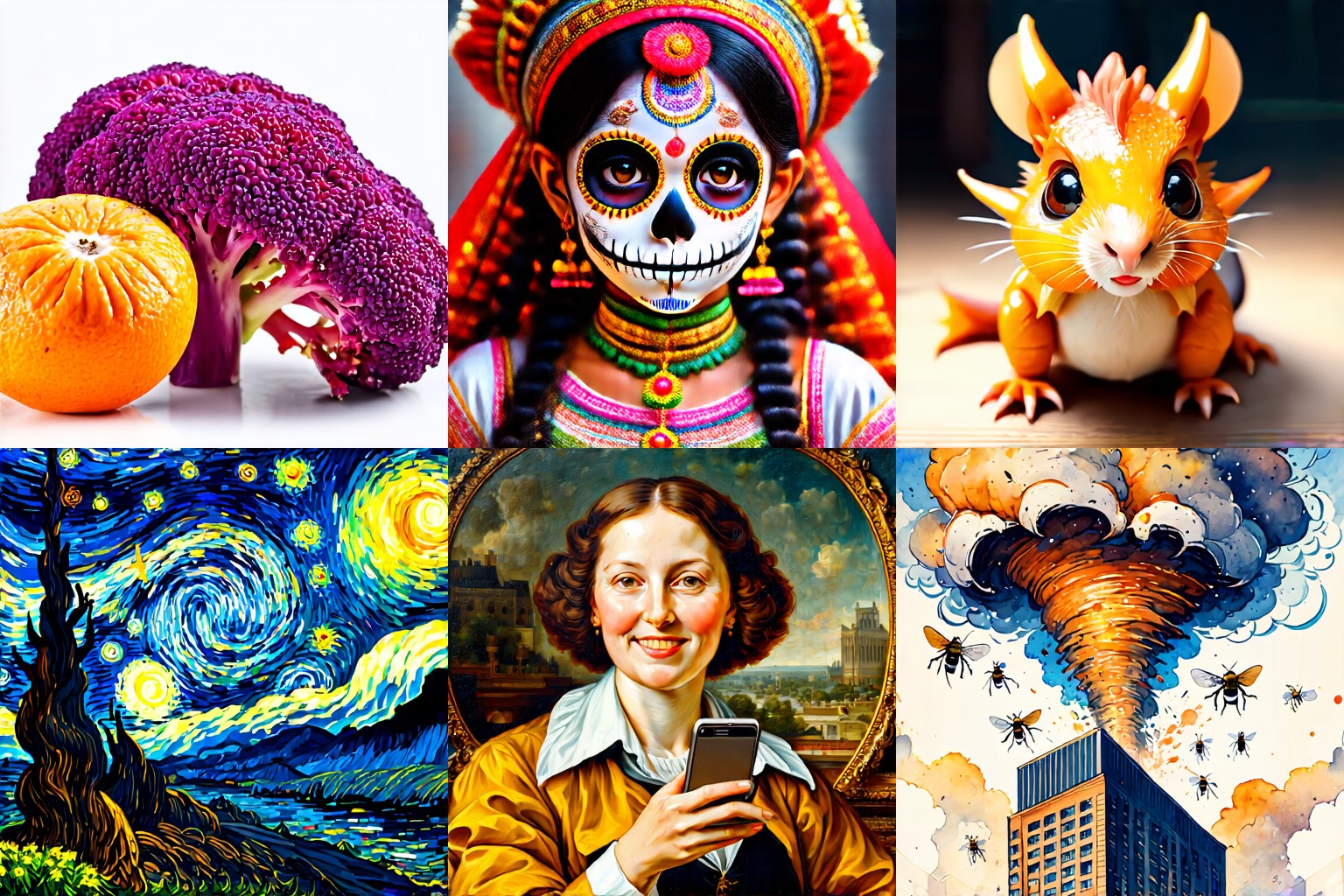}
    \caption{Images generated using SD3.5 with our Quality-Aware Reward Modulation (QRM).
QRM modifies only the AdaLN modulation offsets but produces significantly improved prompt fidelity and visual detail.}
\label{fig:sample images}
\end{figure*}

The field of text-to-image generation is currently dominated by diffusion models \cite{ddpm} such as DALL·E \cite{dalle3}, Midjourney \cite{midjourney}, and Stable Diffusion \cite{stablediffusion}, which are capable of producing high-quality prompt-aligned images. These models generate outputs by iteratively denoising latent noise over a sequence of diffusion timesteps, gradually transforming noise into a structured image. Crucial to text-to-image generation, diffusion models condition the denoising process on textual prompts, allowing the generated image to reflect the semantic content of the input description.

Recent advances have introduced transformer-based diffusion architectures known as Diffusion Transformers (DiT) \cite{dit}, which replace the convolutional U-Net backbone with transformer blocks. These models incorporate conditioning signals at every diffusion timestep that modulate the internal activations of the network. Modern systems such as Stable Diffusion 3.5 (SD3.5) \cite{sd3.5} adopt this architecture, using learned timestep embeddings and pooled prompt embeddings to control the denoising process. While these conditioning signals communicate information about the prompt and the current noise level of the latent image, they do not provide any indication of the quality or fidelity of the intermediate generated image.

In diffusion transformers, this conditioning is applied through modulation layers \cite{dit}. At each diffusion timestep, the conditioning signals are transformed into modulation parameters that scale, shift, and gate the activations of transformer blocks. These parameters adjust the strength of attention and feedforward computations, allowing the model to adapt its denoising behavior based on the prompt and timestep. While this mechanism significantly improves generation quality and stability, the conditioning signals themselves are limited to prompt semantics and timestep information.

This work explores whether modulation can be enhanced beyond timestep and prompt embeddings by incorporating a representation of latent image quality. During the diffusion process, visual fidelity and prompt alignment may degrade due to error accumulation, incorrect prompt interpretation, or hallucinated features \cite{imagereward}. Because the baseline conditioning signals depend only on the timestep and prompt, the modulation mechanism has no way to directly correct these issues based on the evolving state of the latent image.

To address this limitation, we introduce the Quality Representation Module (QRM), a lightweight transformer module \cite{attention} that learns a quality-aware representation during diffusion. The QRM receives the baseline modulation inputs used by diffusion transformers (timestep and pooled prompt embeddings), along with latent image features and the baseline modulation parameters $M_{base}$. Using these inputs, the QRM predicts a set of modulation updates $M_{qrm}$ that refine the baseline modulation parameters at each diffusion timestep. The QRM is trained using a Reward Feedback Learning (ReFL) strategy \cite{imagereward}, which allows the module to learn how modulation changes influence the semantic alignment between the generated image and its associated prompt.

The predicted vectors $M_{qrm}$ are combined with the baseline modulation parameters $M_{base}$ to form the final modulation parameters applied within the transformer blocks. By injecting quality-aware information directly into the modulation pathway, the QRM provides a mechanism for adjusting model activations during the denoising process without modifying the underlying diffusion backbone.

Although the proposed method is designed for diffusion transformer architectures in general, we evaluate QRM using Stable Diffusion 3.5 (SD3.5) \cite{sd3.5} as a representative DiT-based model. This allows us to study the effects of quality-aware modulation within a modern large-scale diffusion transformer while keeping the pretrained backbone weights frozen and the sampling schedule unchanged.

Our contributions are as follows:

\begin{itemize}
\item We propose the Quality Representation Module (QRM), a lightweight transformer-based module \cite{attention} that injects quality-aware modulation into diffusion transformer architectures.
\item We introduce a mechanism for augmenting diffusion transformer modulation layers with latent image quality signals without modifying the underlying diffusion backbone or sampling schedule.
\item We demonstrate that the predicted modulation updates $M_{qrm}$ refine the denoising process at each timestep, improving prompt adherence and image fidelity.
\item We perform extensive evaluations and ablations using image evaluation metrics including CLIP Score \cite{clipscore}, ImageReward \cite{imagereward}, and Human Preference Score (HPS) v2.1 \cite{hpsv2} to analyze the effects of QRM architecture, input features, and training losses.
\end{itemize}

\section{Related Works}

\subsection{Diffusion Models for Text-to-Image Generation}

Diffusion models generate data by progressively transforming Gaussian noise into structured samples through an iterative denoising process~\cite{ddpm}.  
In text-to-image generation, this denoising trajectory is conditioned on textual prompts, enabling the model to synthesize images that reflect the semantic content of the input description.

Latent diffusion models (LDMs)~\cite{ldm} significantly improved the efficiency of diffusion-based generation by performing the diffusion process in a learned latent space rather than directly in pixel space.  
This approach uses a pretrained variational autoencoder (VAE) to encode images into compact latent representations, reducing computational cost while preserving semantic structure.  
Stable Diffusion~\cite{ldm,sdv2,sdxl} popularized this paradigm and demonstrated that high-quality images can be generated by conditioning latent diffusion models on text embeddings.

More recent systems such as Stable Diffusion 3.5 (SD3.5)~\cite{sd3.5} and FLUX~\cite{flux} further advance this framework by replacing the traditional convolutional U-Net backbone with Diffusion Transformers (DiT)~\cite{dit}.  
These architectures use transformer blocks with adaptive normalization layers to incorporate conditioning signals throughout the denoising process, enabling more flexible and scalable generative models.

\subsection{Diffusion Transformer Architecture}

Diffusion Transformer (DiT) architectures~\cite{dit} replace the convolutional U-Net backbone traditionally used in diffusion models with transformer blocks based on the self-attention architecture introduced in~\cite{attention}.  
Similar to earlier latent diffusion systems, these models typically operate in the latent space of a pretrained variational autoencoder (VAE)~\cite{vae}, allowing diffusion to be performed on compact semantic representations rather than full-resolution images.

A key component of DiT models is the incorporation of conditioning signals at every denoising step through adaptive normalization mechanisms.  
Conditioning information is typically derived from two sources: a pooled prompt embedding $p$, produced by pretrained text encoders such as CLIP~\cite{clip} or T5~\cite{t5}, and a timestep embedding $t$, obtained from a sinusoidal encoding of the current diffusion step.  
These embeddings are projected through learned multilayer perceptrons and combined to form a conditioning vector

\begin{equation}
c_t = \mathrm{MLP}_p(p) + \mathrm{MLP}_t(t).
\end{equation}

The conditioning vector $c_t$ is then transformed through learned linear projections to generate the modulation parameters used by Adaptive Layer Normalization (AdaLN)~\cite{dit} within each transformer block.  
For each block, these projections produce vectors corresponding to scale ($\gamma$), shift ($\beta$), and gating terms that modulate the normalized hidden activations.  
The scale and shift parameters adjust the normalized features, while the gating terms control the strength of the attention and feedforward residual pathways.

Thus, the baseline modulation parameters can be viewed as a function of the prompt and timestep embeddings,

\begin{equation}
M_{\text{base}} = f_{\theta}(p,t),
\end{equation}

where $f_{\theta}$ denotes the learned projections that generate AdaLN parameters for each transformer block.

Through this mechanism, prompt semantics and timestep information determine the modulation vectors that govern the behavior of each transformer block during the denoising process.  
These vectors therefore serve as the primary conditioning interface between the prompt, the diffusion timestep, and the internal transformer activations.

In this work we investigate how these modulation parameters can be augmented with additional quality-aware information.  
For our experiments we use Stable Diffusion 3.5 (SD3.5)~\cite{sd3.5}, a large-scale diffusion transformer based on the MM-DiT architecture, as a representative implementation of this design.

\subsection{Quality-Aware Representations in Generative Models}

Assessing and improving the perceptual quality of generated images has become an important challenge in modern text-to-image generation.  
Because human judgments of image quality are difficult to encode directly in training objectives, recent work has increasingly relied on learned reward models that estimate semantic alignment and perceptual fidelity.

Contrastive Language–Image Pre-training (CLIP)~\cite{clip} learns a shared embedding space for images and text, enabling zero-shot classification and semantic similarity measurement.  
CLIPScore~\cite{clipscore} leverages this representation to evaluate prompt–image alignment via cosine similarity.

Building on CLIP, several reward models have been proposed for evaluating text-to-image generation quality, including ImageReward~\cite{imagereward} and Human Preference Score (HPSv2)~\cite{hps,hpsv2}.  
These models extend CLIP-based architectures using human preference datasets to better capture perceptual quality and prompt relevance.

Recent works incorporate such reward signals into diffusion models.  
Reward Feedback Learning (ReFL)~\cite{imagereward} uses a reward model during training, Reward-Instruct~\cite{rewardinstruct} fine-tunes diffusion models using evaluation scores, and Schedule-on-the-Fly~\cite{sonf} dynamically modifies the denoising schedule during inference.

To our knowledge, no prior work integrates quality-aware representations directly into the per-timestep modulation of diffusion transformers.  
Our QRM introduces such a mechanism by predicting additive updates to the AdaLN modulation parameters.

\section{Methodology}
We introduce the Quality-Aware Reward Modulation (QRM), a lightweight module that predicts additive corrections to the baseline AdaLN modulation parameters of a diffusion transformer (instantiated here with SD3.5 \cite{sd3.5}).  
Unlike LoRA-style fine-tuning, QRM does not modify backbone weights; it predicts per-sample, per-timestep modulation offsets that are added to existing AdaLN parameters.  
By injecting quality-aware information into the per-timestep conditioning pathway, QRM enables reward-guided modulation while keeping the diffusion backbone and sampling schedule frozen.
\begin{figure*}[t]
    \centering
    \includegraphics[width=.90\textwidth]{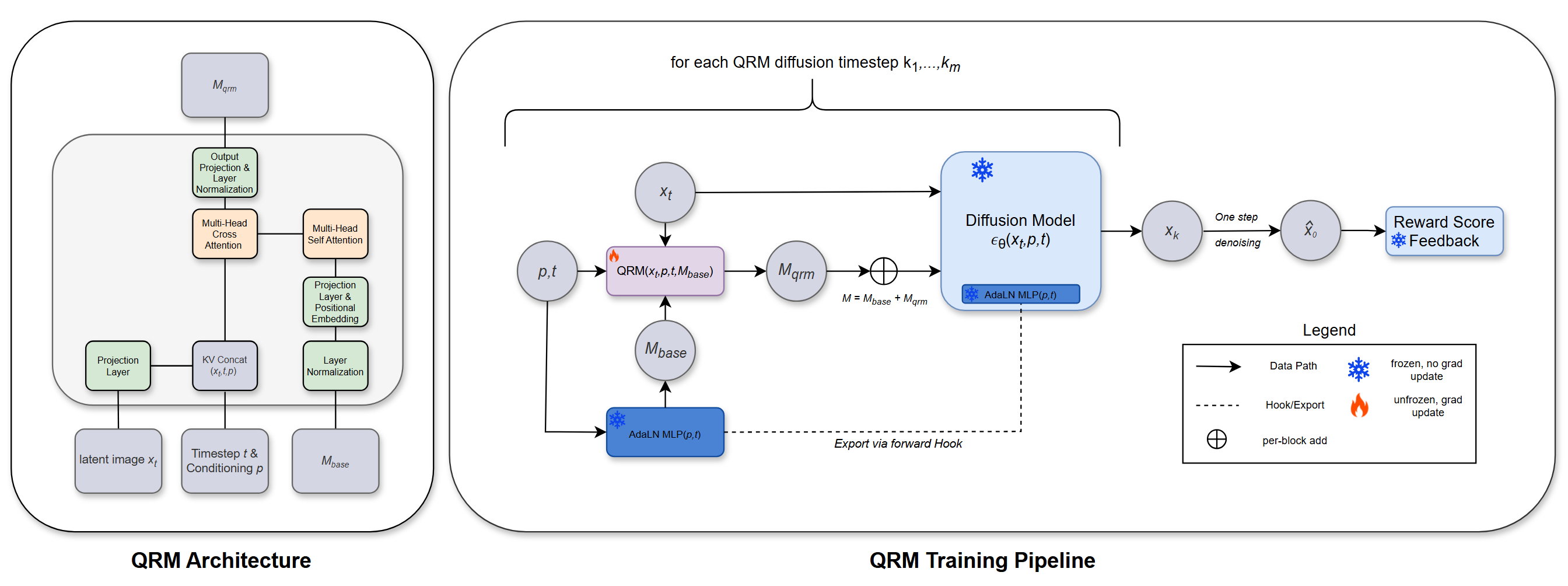}
    \caption{Overview of the proposed QRM. Left: The QRM architecture is a transformer decoder that combines the latent image $x_t$, timestep $t$, prompt embedding $p$, and baseline modulation $M_{\text{base}}$ 
to produce modulation updates $M_{\text{qrm}}$. Right: In training, diffusion is applied from timestep $T$ to $k$. At each diffusion timestep, the baseline modulation $M_{\text{base}}$ from the diffusion model’s AdaLN layers is combined with $M_{\text{qrm}}$ to produce updated modulation vectors. The diffusion model eventually generates an latent image $x_k$, which produces $\hat{x}_0$ after one-step denoising. The predicted $\hat{x}_0$ estimate is then evaluated by a frozen reward model to provide a scalar feedback score.}
    \label{fig:qrm}
\end{figure*}

\subsection{Why latent-dependent modulation helps.}
In diffusion transformers \cite{dit}, AdaLN modulation parameters are typically computed only from timestep and prompt conditioning. As a result, the baseline modulation for a given prompt--timestep pair is independent of the current latent state:
\begin{equation}
M_{\text{base}} = f_{\theta}(p,t),
\qquad
\frac{\partial M_{\text{base}}}{\partial x_t} = 0.
\end{equation}
This prevents the modulation pathway from directly reacting to sample-specific failure modes that emerge in the evolving latent (e.g., missing attributes or hallucinated relations). QRM introduces a latent-dependent correction
\begin{equation}
M = M_{\text{base}} + M_{\text{qrm}}, 
\qquad 
M_{\text{qrm}} = f_{\text{QRM}}(x_t, M_{\text{base}}, p, t),
\label{qrm_forward}
\end{equation}
which creates a lightweight feedback channel from the current latent $x_t$ to the AdaLN parameters without modifying the denoiser or sampler.

\subsection{QRM Architecture}
The proposed QRM augments the pretrained diffusion backbone by generating a modulation update $M_{\text{qrm}}$ that refines the baseline modulation parameters $M_{\text{base}}$ within the diffusion transformer~\cite{dit}.

The QRM is implemented using a transformer decoder architecture~\cite{attention}: its cross-attention layers allow modulation query tokens to attend over a fixed key--value memory composed of latent-image tokens, pooled prompt embeddings, and timestep embeddings. It receives as input: (i) the latent image $x_t$ at the current denoising step, (ii) the pooled prompt embedding $p$, (iii) the timestep embedding $t$, and (iv) the baseline modulation parameters $M_{\text{base}}$, which consist of concatenated scale, shift, and gate vectors extracted from each AdaLN layer of the diffusion model~\cite{dit}. Each contiguous segment of this concatenated vector is referred to as a span, corresponding to a single AdaLN layer.

From the current latent image $x_t$ we form $P=S^2$ latent tokens by projecting each spatial location to a width $d$ through a learned $1{\times}1$ convolution $W_\ell$, and then apply an adaptive pooling operation to obtain a $S{\times}S$ grid of latent representations:
\begin{equation}
X=\operatorname{flat}\!\big(\operatorname{Pool}_{S\times S}(W_\ell^{1\times1} * x_t)\big).
\end{equation}
For each span $i$, we build one \emph{modulation query} token by normalizing and projecting a slice of the modulation parameters $u_i$ that corresponds with the $i^{th}$ layer, adding a learned bias $b_i^{in}$, then adding a learned positional embedding $\Pi$:
\begin{equation}
z_i=W^{\text{in}}_i\,\operatorname{LN}(u_i)+b^{\text{in}}_i,
\end{equation}
\begin{equation}
Z_0=\big[z_1;\ldots;z_M\big]+\Pi.
\end{equation}
Two additional conditioning tokens are obtained from the pooled prompt and timestep via a single linear projection to width $d$:
\begin{equation}
k_y=W_y\,\operatorname{LN}(p),\qquad
k_t=W_t\,t,
\end{equation}
and we form the key--value memory
\begin{equation}
\mathrm{KV}=\big[X\;\|\;k_y\;\|\;k_t\big].
\end{equation}

${Z} \in \mathbb{R}^{M \times d}$ denote the set of query tokens that attend to ${KV}$ through a stack of $L$ pre-norm transformer decoder layers~\cite{attention}:
\begin{align*}
Z &\leftarrow Z + \mathrm{MSA}\!\big(\operatorname{LN}(Z)\big),\\
Z &\leftarrow Z + \mathrm{MHA}\!\big(\operatorname{LN}(Z),\,\operatorname{LN}(\mathrm{KV}),\,\operatorname{LN}(\mathrm{KV})\big),\\
Z &\leftarrow Z + \mathrm{FF}\!\big(\operatorname{LN}(Z)\big).
\end{align*}
A final LayerNorm gives $Z_{\text{out}}=\operatorname{LN}(Z)$, which is projected back to the native AdaLN widths per span and concatenated to form the offset:
\begin{equation}
M_{\text{qrm}}
= \operatorname*{concat}_{i=1}^{M}
\big( W^{\text{out}}_i\, Z_{\text{out},i} + b^{\text{out}}_i \big),
\end{equation}
where $Z_{\text{out},i}$ denotes the token associated with span $i$.

The corrected modulation vectors are obtained by $M = M_{\text{base}} + M_{\text{qrm}}$. As in the baseline architecture~\cite{dit}, the resulting modulation provides scale ($\gamma$), shift ($\beta$), and gate ($g$) parameters for AdaLN within each transformer block. In all experiments, the diffusion backbone (denoiser, text encoders, timestep embedding, VAE, and sampler) remains frozen; only QRM parameters are optimized.

\subsection{Training Methodology}
During training, gradients update only QRM parameters; the denoiser, text encoders, VAE, and reward model remain frozen.

We adopt a reward--feedback learning scheme following ReFL~\cite{imagereward}: the diffusion backbone and reward model are frozen, and only the QRM modulator is updated to increase a scalar reward for an image decoded from an intermediate denoising step.

Let $\{\sigma_t\}_{t=1}^T$ denote the baseline diffusion schedule with corresponding latent states $x_t$.
The model follows this schedule for all diffusion steps, while QRM is activated only on a fixed subset of indices $\mathcal{K}_{\text{qrm}}\subseteq\{1,\dots,T\}$.
This restriction improves efficiency and concentrates QRM capacity on the timesteps that most strongly influence global structure and semantic content (early and mid-stage denoising), while later steps primarily refine high-frequency details.

For each active step $k \in \mathcal{K}_{\text{qrm}}$, the backbone exposes the AdaLN modulation $M_{\text{base},k}$, and QRM predicts an offset:
\begin{equation}
M_{\text{qrm},k} = f_{\text{QRM}}(x_k, M_{\text{base},k}, p, t_k).
\end{equation}
which is then combined with the baseline modulation:
\begin{equation}
M_k = M_{\text{base},k} + M_{\text{qrm},k}.
\end{equation}

During training, an index $\hat{k}$ is sampled from $\mathcal{K}_{\text{qrm}}$.
The baseline diffusion process proceeds normally, and QRM modulation is applied only when the current step $k \in \mathcal{K}_{\text{qrm}}$.
Upon reaching $\hat{k}$, a single denoising update is performed using the current latent $x_{\hat{k}}$ and the QRM-modulated model output.
Let $\epsilon_{\theta}$ denote the frozen denoising network and $\Phi$ the sampler update rule (e.g., DPM++(2M)). We write $\epsilon_\theta(\cdot)$ as consuming the effective modulation $M_{\hat{k}}$ (derived from $(p,t_{\hat{k}})$), suppressing other conditioning arguments for brevity.
The one-step latent update is:
\begin{equation}
\tilde{x}_{\hat{k}} = \epsilon_{\theta}(x_{\hat{k}}, M_{\hat{k}}, p, t_{\hat{k}}),
\end{equation}
\begin{equation}
x_{\hat{k}-1} = \Phi(x_{\hat{k}}, \tilde{x}_{\hat{k}}, \sigma_{\hat{k}}).
\end{equation}
The updated latent is decoded by the frozen VAE decoder $D$:
\begin{equation}
I_{\hat{k}-1} = D(x_{\hat{k}-1}),
\end{equation}
and scored by a frozen reward model $S(\cdot,p)$ (e.g., HPSv2~\cite{hpsv2} or CLIPScore~\cite{clipscore}):
\begin{equation}
s_{\hat{k}-1} = S(I_{\hat{k}-1}, p).
\end{equation}

Gradients flow only through the QRM-modified branch; the backbone and reward model remain frozen. We use one of the following ReFL-style losses~\cite{imagereward}.

\emph{Reward-maximization loss:}
\begin{equation}
\mathcal{L}_{\text{reward}}
=
\mathbb{E}_{(x,p),\,k\sim\mathcal{I}}
\bigl[\, m - s_{\hat{k}-1}^{\text{qrm}} \,\bigr]_+
\end{equation}

\emph{Margin-seeking loss:}
\begin{equation}
\mathcal{L}_{\text{margin}}
=
\mathbb{E}_{(x,p),\,k\sim\mathcal{I}}
\bigl[\, m - ({s_{\hat{k}-1}^{\text{qrm}}} - {s_{\hat{k}-1}^{\text{base}}}) \,\bigr]_+,
\end{equation}
which pushes the QRM image to outperform the baseline by a margin $m>0$. The baseline score $s_{\hat{k}-1}^{\text{base}}$ is obtained using the same procedure but with $M_{\hat{k}}=M_{\text{base},\hat{k}}$.

\subsection{QRM at Inference}

At inference time, the trained QRM module is loaded alongside the pretrained diffusion backbone and remains frozen.  
Given the current latent state $x_t$, pooled prompt embedding $p$, timestep embedding $t$, and baseline modulation parameters $M_{\text{base},t}$ exposed by the diffusion transformer\cite{dit}, QRM predicts a modulation offset
\begin{equation}
M_{\text{qrm},t} = f_{\text{QRM}}(x_t, M_{\text{base},t}, p, t),
\end{equation}
as defined in Eq.~\ref{qrm_forward}.  
The final modulation applied inside each transformer block is then obtained by
\begin{equation}
M_t = M_{\text{base},t} + M_{\text{qrm},t}.
\end{equation}

These updated modulation parameters replace the baseline AdaLN vectors used by the diffusion transformer while leaving all backbone weights unchanged.  
As a result, QRM acts as a lightweight quality-aware adapter that modifies the denoising computation through conditioning rather than parameter updates.

During sampling, QRM is applied only on a predefined subset of diffusion timesteps $\mathcal{K}_{\text{qrm}}$.  
For timesteps $t \in \mathcal{K}_{\text{qrm}}$, the denoiser prediction incorporates the QRM-modulated conditioning, while for $t \notin \mathcal{K}_{\text{qrm}}$ the model reduces to the baseline update using $M_{\text{base},t}$.  
The sampler update therefore takes the form
\begin{equation}
x_{t-1} = \Phi\!\left(x_t,\,\epsilon_{\theta}(x_t, M_t, p, t),\,\sigma_t\right),
\end{equation}
where $\epsilon_{\theta}$ denotes the frozen denoising network and $\Phi$ represents the sampler update rule(e.g., DPM-Solver++(2M)\cite{dpm_solver}).

Restricting QRM to a subset of timesteps serves two purposes.  
First, it reduces computational overhead by avoiding unnecessary modulation during stages where the latent trajectory is already well determined.  
Second, it concentrates QRM’s influence on diffusion steps where global structure and semantic alignment are most sensitive to modulation, particularly during the high-noise portion of the denoising process.

Because QRM operates entirely through additive modulation offsets, the inference procedure remains fully compatible with existing diffusion pipelines.  
No retraining or architectural modifications of the backbone are required, and the additional computation introduced by QRM is small relative to the cost of the denoiser itself.  
In practice, QRM therefore provides a lightweight mechanism for steering the denoising trajectory toward latent states that better reflect the reward-informed quality signals learned during training while preserving the baseline model behavior outside the active timestep window.
\section{Experiments}
\subsection{Experimental Setup}

\paragraph{Dataset.}
We train using the ImageReward prompt set~\cite{imagereward}, which contains 10{,}000 diverse text prompts. 
For evaluation we use Parti-Prompts~\cite{parti}, a standard benchmark of 1{,}632 prompts covering a wide range of compositional and attribute-based instructions.

\paragraph{Implementation Details.}
We use the open-source Stable Diffusion 3.5 Medium~\cite{sd3.5} as the frozen diffusion backbone. 
QRM is trained with Adam at a learning rate of $2.5\times10^{-5}$ using gradient accumulation of 3 and batch size 1. 
Training runs for 25 epochs with 399 optimization steps per epoch; each step processes one prompt sampled uniformly from the ImageReward set, resulting in 9{,}975 total updates (approximately one pass over the 10{,}000 training prompts).
We use the default SD3.5 sampler, DPM-Solver++(2M)~\cite{sd3.5}. 
Unless otherwise stated, QRM is activated only during the early (high-noise) denoising timesteps $t\in\{50,\ldots,26\}$ of the diffusion schedule ($T=50$ in our setup).
The diffusion backbone and reward model are frozen; gradients update only QRM. 
All experiments were run on an RTX 5090.

\paragraph{Reward Models and Evaluation Metrics.}
During training we test CLIPScore~\cite{clipscore} and HPSv2.1~\cite{hpsv2} as frozen reward scorers.
For evaluation we report HPSv2.1~\cite{hpsv2} and an Aesthetic Predictor~\cite{aesthetic} as primary metrics, and additionally use CLIPScore~\cite{clipscore} and ImageReward~\cite{imagereward} for ablations and analysis.

\paragraph{Evaluation Protocol and Fairness.}
All QRM comparisons use the \emph{same} SD3.5 backbone, sampler, guidance settings, prompt sets, and random seeds as the baseline; the only change is adding $M_{\mathrm{qrm}}$ to the AdaLN modulation on the chosen timestep subset $\mathcal{K}_{\mathrm{qrm}}$.
Unless otherwise stated, we generate one image per prompt on Parti-Prompts using identical inference hyperparameters for baseline and QRM.
We report dataset-level averages of each metric, and qualitative figures use matched seeds to visualize how QRM changes the denoising trajectory rather than selecting favorable samples.

\subsection{Quantitative Comparison}
\begin{table}[!h]
\centering
\small
\setlength{\tabcolsep}{1pt}
\caption{Comparison on the Parti-Prompts benchmark~\cite{parti}. Results for SD1.5-based prompt refinement, diffusion fine-tuning, and ReNeg methods are reported from the ReNeg benchmark~\cite{reneg} for reference and are not directly comparable to SD3.5. Our comparison focuses on improvements to the SD3.5 backbone (baseline vs.\ SD3.5+QRM) under the same inference setup. Bold values denote the best per metric.}
\label{tab:parti_prompts_comparison}
\begin{tabular}{l l c c}
\toprule
\textbf{Category} & \textbf{Model} & \textbf{Aesthetic} $\uparrow$ & \textbf{HPSv2.1} $\uparrow$ \\
\midrule

\multirow{2}{*}{Direct Inference} 
 & SD1.5                & 5.23 & 25.67 \\
 & SD3.5        & 6.02 & 29.15 \\
\midrule

\multirow{3}{*}{Prompt Refinement}
 & BeautifulPrompt & 5.78 & 22.72 \\
 & Promptist      & 5.42            & 25.24 \\
 & DNP   & 5.21            & 25.83 \\
\midrule

\multirow{5}{*}{Finetuning SD}
 & DDPO-Aesthetic & 4.99 & 20.69 \\
 & DDPO-Alignment  & 4.93 & 19.00 \\
 & Diffusion-DPO & 5.26 & 26.62 \\
 & ReFL & 5.48 & 27.97 \\
 & TextCraftor (Text) & 5.90 & 28.36 \\
\midrule

\multirow{2}{*}{ReNeg}
 & Global Neg. Emb.     & 5.45 & 29.16 \\
 & Per-sample Neg. Emb. & 5.50 & 29.84 \\
\midrule

 \multirow{2}{*}{QRM}
 & SD3.5 (CLIP) & 6.04 & 29.46 \\
 & SD3.5 (HPSv2.1)     & \textbf{6.22} & \textbf{29.97} \\
\bottomrule
\end{tabular}
\end{table}

To position QRM relative to existing post-training and reward-guided diffusion methods, we compare against a range of established baselines on the Parti-Prompts benchmark~\cite{parti}. Baseline results for SD1.5, prompt-refinement methods, diffusion finetuning techniques, and ReNeg are reported directly from~\cite{reneg}. 

It should be noted that Prompt Refinement, Finetuning SD, and ReNeg methods were originally evaluated using SD1.5. Our goal is therefore not to compare SD3.5 directly against SD1.5-based systems, but to show that QRM further improves an already strong diffusion model without modifying its backbone weights. All methods are evaluated using the Aesthetic~\cite{aesthetic} and HPSv2.1~\cite{hpsv2} metrics, which measure perceptual quality and human-preference alignment, respectively.

The QRM-enhanced SD3.5 model achieves the best performance on both Aesthetic and HPSv2.1 metrics despite modifying only the AdaLN modulation parameters while keeping the SD3.5 backbone frozen. When trained using CLIPScore as the reward signal, QRM improves human-preference alignment over the SD3.5 baseline while maintaining comparable perceptual quality. 

More notably, training QRM with the HPSv2.1 reward metric yields the highest Aesthetic and HPSv2.1 scores among all evaluated methods. Compared to the SD3.5 baseline, QRM improves Aesthetic by +0.20 and HPSv2.1 by +0.82 on the Parti-Prompts benchmark. These results demonstrate that QRM can consistently improve human-preference alignment for diffusion transformers while leaving the pretrained backbone entirely frozen and without introducing additional LoRA-style finetuning.

\subsection{Qualitative Demonstration}

\begin{figure*}[!t]
    \centering
    \includegraphics[width=\textwidth]{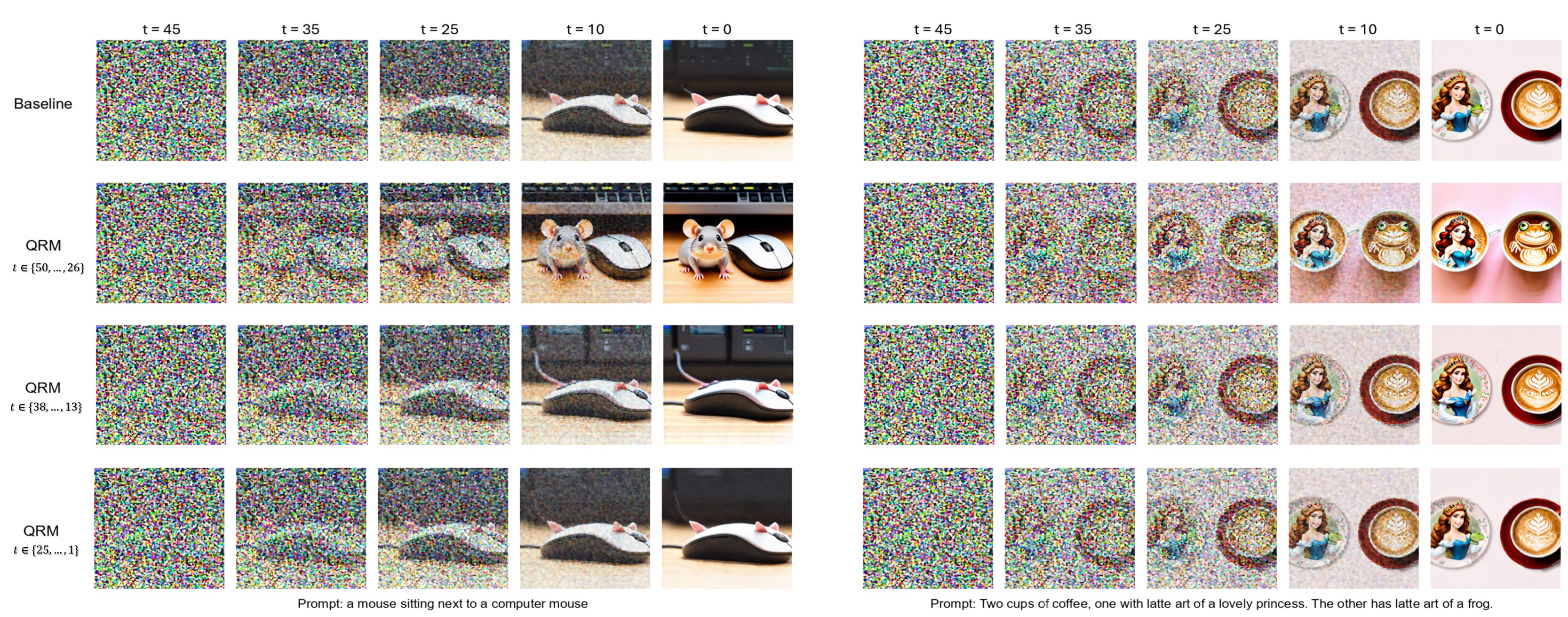}
    \caption[Qualitative Demonstration: Visualizing Denoising Trajectories]{Denoising trajectories for sample prompts.
Top row: baseline SD3.5; Bottom 3 rows: SD3.5 + QRM with different timestep ranges.
Columns show the decoded latent image at diffusion steps $t\in\{45,35,25,10,0\}$.}
\label{fig:qualitative}
\end{figure*}

To illustrate how QRM affects the denoising trajectory, 
Figure~\ref{fig:qualitative} visualizes intermediate latent reconstructions at several diffusion steps. 
All trajectories are generated from the same initial noise seed to enable direct comparison between the baseline and QRM-modulated runs.

The top row shows the baseline SD3.5 trajectory, while the remaining rows apply QRM during different portions of the denoising schedule: \textbf{early ($t\!\in\!\{50,\ldots,26\}$)}, \textbf{middle ($t\!\in\!\{38,\ldots,13\}$)}, and \textbf{late ($t\!\in\!\{25,\ldots,1\}$)} timesteps.

When QRM is applied during the early denoising stages, recognizable semantic structure emerges significantly earlier in the trajectory. 
Objects become well-formed by $t=25$, and the final images show stronger adherence to the prompt. 
In contrast, applying QRM only during the middle or late stages produces noticeably smaller improvements, as much of the global structure has already been determined by that point.

These results support the intuition that QRM is most effective when applied during high-noise timesteps where global structure and semantic alignment are first established. 
Once the denoising process reaches later stages, corrections to the latent representation become more limited, resulting in smaller visual differences relative to the baseline.

\subsection{Ablation Study on QRM Design Choices}

We conduct ablations to evaluate the effect of key QRM design choices on image quality and alignment performance.
Each variant is trained under identical data and optimization settings while varying only
(i) the \textbf{QRM architecture},
(ii) the \textbf{active timestep range} $\mathcal{K}_{\text{qrm}}$, and
(iii) the \textbf{training reward metric}.
All models are trained on the same subset of the dataset and evaluated on the Parti-Prompts benchmark~\cite{parti}
using CLIPScore~\cite{clipscore}, ImageReward~\cite{imagereward}, and HPSv2.1~\cite{hpsv2}.

\paragraph{Architectural Variants.}

We compare three QRM architectures of increasing capacity: 
\textit{small}, \textit{medium}, and \textit{large}.
These correspond to transformer configurations $(L,d)=(4,384)$, $(6,512)$, and $(8,640)$ respectively,
where $L$ is the decoder depth and $d$ is the token dimension.

Table~\ref{tab:qrm_ablation_architecture} shows that larger QRM variants achieve higher HPSv2.1 scores but slightly lower CLIPScore and ImageReward values.
Because the training objective uses HPSv2.1 as the reward signal, increasing model capacity allows QRM to more strongly optimize toward this metric,
potentially at the expense of other semantic similarity metrics.

\begin{table}[h]
\centering
\caption{Ablation on QRM model size. Each model is trained with HPSv2.1 as the reward loss metric and a fixed QRM timestep range $\mathcal{K}_{\text{qrm}} = \{35,\ldots,25\}$ for modulation (timestep effects are analyzed separately in Table~\ref{tab:qrm_ablation_timesteps}). The model dimension refers to the feature vector length per token. Bold results indicate the highest score for that metric.}
\label{tab:qrm_ablation_architecture}
\begin{tabular}{lccc}
\toprule
\textbf{Model Size (dim)}  & \textbf{CLIP} $\uparrow$ & \textbf{IR} $\uparrow$ & \textbf{HPSv2.1} $\uparrow$ \\
\midrule
Small (384) & \textbf{32.97} & \textbf{1.19} & 29.11 \\
Medium (512) & 32.81 & 1.17 & 29.19 \\
Large (640) & 32.71 & 1.16 & \textbf{29.30} \\
\bottomrule
\end{tabular}
\end{table}

\paragraph{Timestep Range.}

To examine when QRM modulation is most effective, we vary the active diffusion range $\mathcal{K}_{\text{qrm}}$ while keeping all other settings fixed.
Table~\ref{tab:qrm_ablation_timesteps} shows that applying QRM during early denoising stages ($\{50,\ldots,26\}$) yields the highest HPSv2.1 score.

In contrast, applying QRM only during later timesteps ($\{25,\ldots,1\}$) produces noticeably weaker alignment improvements.
This suggests that reward-guided modulation is most effective when injected during the early structural phase of the diffusion process,
when global semantic structure is first established.

\begin{table}[h]
\centering
\caption{Ablation on the active timestep range $\mathcal{K}_{\text{qrm}}$. Results are for Large (640) model size QRM trained with HPSv2 reward metric. Bold results indicate the highest score for that metric.}
\label{tab:qrm_ablation_timesteps}
\begin{tabular}{lccc}
\toprule
\textbf{Timestep Range} & \textbf{CLIP} $\uparrow$ & \textbf{IR} $\uparrow$ & \textbf{HPSv2.1} $\uparrow$ \\
\midrule
$\{50,\ldots,26\}$   & 32.50 & \textbf{1.18} & \textbf{29.97} \\
$\{38,\ldots,13\}$  & 32.93 & \textbf{1.18} & 29.38 \\
$\{25,\ldots,1\}$  & \textbf{33.17} & 1.15 & 27.75 \\
\bottomrule
\end{tabular}
\end{table}

\paragraph{Training Reward Metric.}

Finally, we compare QRM models trained with either CLIPScore or HPSv2.1 as the optimization signal under the same margin-seeking loss.
Both reward models aim to improve image–text alignment but emphasize different aspects of visual quality.

As shown in Table~\ref{tab:qrm_ablation_loss}, training with CLIPScore produces higher CLIPScore and ImageReward values,
while training with HPSv2.1 yields higher HPSv2.1 scores.
This indicates that the reward model used during training directly shapes the perceptual biases learned by QRM.

\begin{table}[h]
\centering
\caption{Comparison of reward metric with Large (640) model size QRM with $\mathcal{K}_{\text{qrm}} = \{50,\ldots,26\}$. Bold results indicate the highest score for that metric.}
\label{tab:qrm_ablation_loss}
\begin{tabular}{lccc}
\toprule
\textbf{Reward Metric} & \textbf{CLIP} $\uparrow$ & \textbf{IR} $\uparrow$ & \textbf{HPSv2.1} $\uparrow$ \\
\midrule
CLIPScore & \textbf{33.12} & \textbf{1.20} & 29.00 \\
HPSv2.1 & 32.50 & 1.18 & \textbf{29.97} \\
\bottomrule
\end{tabular}
\end{table}

\paragraph{Discussion.}

Across all ablations, three consistent trends emerge.

First, increasing QRM capacity improves alignment with the reward used during training but slightly reduces performance on alternative metrics,
suggesting a degree of reward-specific specialization.

Second, applying QRM during earlier denoising timesteps consistently yields stronger alignment improvements,
supporting the hypothesis that early diffusion stages determine the global semantic structure of the generated image.

Third, the reward metric used for training strongly influences which perceptual characteristics the model learns to emphasize.

Together, these findings suggest that QRM functions as a flexible quality-aware adapter whose behavior can be steered by both the reward signal and the timestep range at which modulation is applied.

\section{Conclusion}

We introduced the Quality Representation Module (QRM), a lightweight transformer module that injects a quality-aware signal into the AdaLN modulation parameters of diffusion transformers. 
Unlike most reward-guided diffusion approaches, QRM improves generation quality without finetuning the diffusion backbone, instead operating entirely through additive modulation updates conditioned on the evolving latent representation.

Experiments on the Parti-Prompts benchmark demonstrate that QRM consistently improves the performance of Stable Diffusion 3.5. 
Compared to the SD3.5 baseline, QRM improves Aesthetic by +0.20 and HPSv2.1 by +0.82 while leaving the pretrained model weights unchanged. 
These results suggest that modulation-level interventions provide an effective mechanism for aligning large diffusion models with human-preference signals.

While promising, QRM inherits the biases of the reward models used during training, and different reward metrics emphasize different notions of image quality. 
Future work may explore multi-reward training, more expressive modulation architectures, and adaptive timestep selection to further improve robustness and generalization. Another promising direction is evaluating QRM on additional diffusion transformer architectures to assess how well the proposed modulation mechanism generalizes beyond SD3.5.
More broadly, our results suggest that targeted modulation of conditioning signals may provide a scalable pathway for aligning diffusion transformers without the need for expensive model retraining.
\bibliographystyle{splncs04}
\bibliography{main}

\newpage
\appendix
\section{Supplemental Information}
\subsection{Baseline Images for Figure~1}

\begin{figure}
\centering
\includegraphics[width=0.95\textwidth]{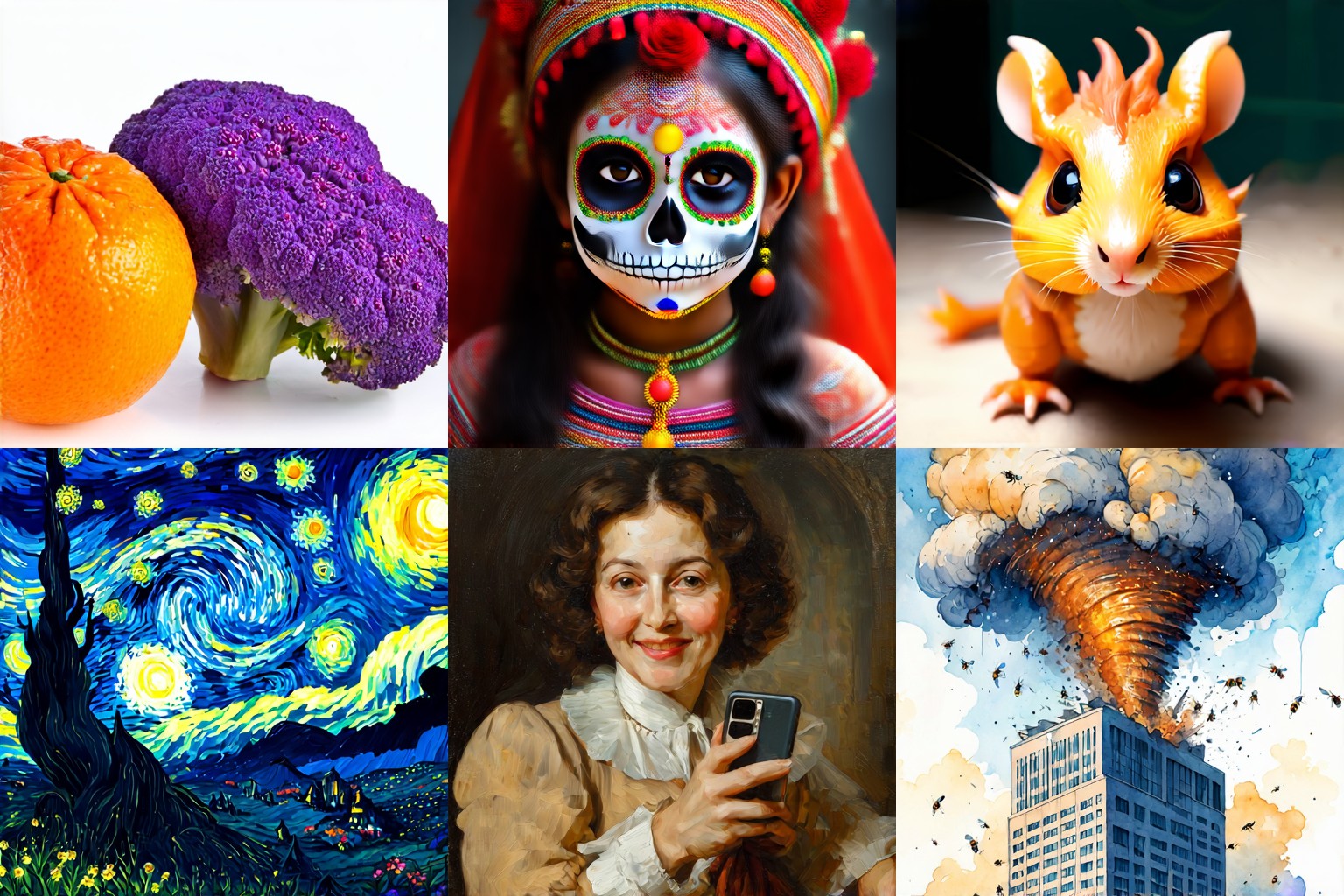}
\caption{Images generated using Baseline SD3.5 (No QRM) corresponding to Figure 1 in the primary text.}
\label{fig:app-baseline-fig21}
\end{figure}
The prompts used for Figure~1 in the primary text are as follows:

\noindent\textbf{First row (left to right):}
\begin{enumerate}
\item ``a photo of a red orange and a purple broccoli''
\item ``ultra detailed, beautiful cute girl wearing modern stylish costume in the style of Assamese bihu mekhela sador gamosa design, dia de los muertos, scifi, cyberpunk, fantasy, intricate details, eerie, movie still, airbrush, elegant, super highly detailed, professional digital painting, artstation, concept art, smooth, sharp focus, no blur, no dof, extreme illustration, Unreal Engine 5, Photorealism, HD quality, 8k resolution, cinema 4d, 3D, beautiful, cinematic, art by artgerm and michael welan and DZO and greg rutkowski and alphonse mucha and loish and WLOP''
\item ``a hamster dragon''
\end{enumerate}

\noindent\textbf{Second row (left to right):}
\begin{enumerate}
\item ``The Starry Night''
\item ``close-up portrait of a smiling businesswoman holding a cell phone, oil painting in the style of Rembrandt''
\item ``A tornado made of bees crashing into a skyscraper. painting in the style of Hokusai.''
\end{enumerate}

\subsection{Computational Efficiency}
\label{sec:efficiency}

Although QRM introduces an additional transformer module into the denoising pipeline, the resulting computational overhead remains moderate in practice. QRM operates only during a subset of denoising steps and processes a compact representation derived from latent and modulation tokens, which limits the additional computational cost.

During inference, enabling QRM increases generation time due to the transformer layers used to compute modulation offsets. For a standard 50-step sampling schedule at a resolution of $512 \times 512$, sampling with the baseline SD3.5 Medium model~\cite{sd3.5,sd3} requires approximately 5 seconds per image, whereas sampling with QRM requires approximately 6 seconds. This corresponds to an overhead of roughly $20\%$. The exact timing depends on the number of denoising steps during which QRM is active and the size of the transformer layers used within the module.

In terms of memory usage, the QRM parameters are stored separately from the SD3.5 backbone and occupy approximately 2.5\,GB when stored in fp32 precision. When enabled during inference, peak VRAM usage increases from roughly 10\,GB for the baseline SD3.5 Medium model to approximately 12.5\,GB. This increase is primarily due to attention buffers and intermediate activations within the QRM transformer layers. The remainder of the diffusion pipeline remains unchanged, as QRM reuses latent and modulation tensors already produced by the underlying model.

Training QRM introduces additional computational overhead. Because QRM is trained using a single randomly selected diffusion timestep per iteration, the dominant additional cost arises from evaluating the QRM transformer layers and computing reward-based quality signals used during optimization. In practice, the per-iteration training time increases by approximately $25$--$30\%$ compared to a training loop without QRM, without requiring additional memory-intensive components beyond the QRM module itself. Training therefore remains feasible on a single modern high-memory GPU.

Overall, QRM introduces moderate computational overhead relative to the underlying diffusion backbone while providing improved image quality and alignment, making it a practical extension to SD3.5-based diffusion pipelines.

\end{document}